# Intellectual Property Evaluation Utilizing Machine Learning

[#]Jinxin Ding
*Department of Economics and Finance*
*City University of Hong Kong*
Hong Kong
jinxiding3-c@my.cityu.edu.hk

[#]Yuxin Huang
*School of Data Science*
*City University of Hong Kong*
Hong Kong
yuxhuang6-c@my.cityu.edu.hk

[#]Keyang Ni
*School of Data Science*
*City University of Hong Kong*
Hong Kong
keyangni2-c@my.cityu.edu.hk

[#]Xueyao Wang
*School of Data Science*
*City University of Hong Kong*
Hong Kong
xueyao.wangwinnie@foxmail.com

[#]Yinxiao Wang
*Department of Computer Science*
*City University of Hong Kong*
Hong Kong
yinxiwang3-c@my.cityu.edu.hk

[#]Yucheng Wang
*College of Business*
*City University of Hong Kong*
Hong Kong
yuchwang8-c@my.cityu.edu.hk

*Abstract*— Intellectual properties is increasingly important in the economic development. To solve the pain points by traditional methods in IP evaluation, we are developing a new technology with machine learning as the core. We have built an online platform and will expand our business in the Greater Bay Area with plans.

## I. BACKGROUND

Intellectual Property is an intangible asset, including copyrights, patents, trademarks, and trade secrets, owned and legally protected by a company or individual from outside use without consent. According to researchers around the world, Intellectual Property incentives finance, creates jobs, optimizes social utility, plays a significant role in the contemporary economy, and the related industries are developing rapidly with huge market.

To realize the full potential of an intangible asset, the first step is IP valuation. IP valuation expresses the contribution of IP to a business in a generally understandable economic value.

The major traditional IP valuation methods are the Present Earning Value Method, Market Comparison Method, and Cost Method. However, due to the special nature of IP itself, and the related regulations are not exhaustive, the traditional IP valuation methods are difficult to use, the valuation time is long, and the cost is high. To better leverage the value of intellectual property, we use machine learning methods to conduct efficient, accurate, and objective valuations.

## II. SOLUTIONS

### A. Detailed Implementation

In order to solve the commercial[1] demand for patents in the market, simplify the complicated steps of analyzing the value of patents, and obtain more accurate analysis results, we propose a Fintech pipeline to analyze the value and quality of patents in the Greater Bay Area (GBA). Our pipeline takes PDF patent documents as input, and the market valuation and quality score of the patent as output, and some basic information of the patent can be extracted, such as applicant, patent validity period.

We utilize machine learning models in our pipeline. Generally speaking, classification, data dimension reduction and regression are applied to the model. For classification, we choose the Support Vector Machine (SVM). In machine learning, Support Vector Machines are supervised learning models with associated learning algorithms that can analyze data for classification (Cortes & Vapnik, 1995). High precision patent classification systems can be developed by using Support Vector Machine (Wu, Ken & Huang, 2010). For the data dimension reduction part, we make use of self-organizing maps (SOMs). Self-organizing maps belong to neural networks. The data are automatically grouped according to the similarities and regular patterns found in the dataset, using some form of distance measurement such as the L2-norm distance. In this way, we can have different data groups differentiated by their quality. For regression, Neural Network is selected. A neural network is a network or circuit of biological neurons, or, a in a modern sense, an artificial neural network, composed of artificial neurons or nodes (Hopfield, 1982). Neural networks can be used in different fields, in this platform, Neural Network is mainly used for regression analysis.

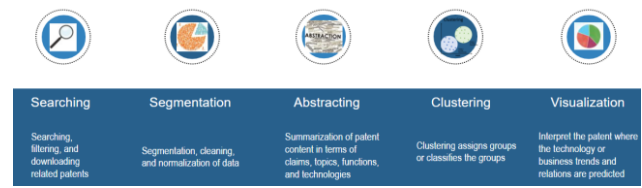

Figure 1 Patent Analysis Pipeline

Figure 1 shows the overall pipeline of the platform. First, searching is applied for searching, filtering, and downloading related patents. Second, we need to do segmentation, which means cleaning, and normalization of structured and unstructured patent data. Third, we use abstracting technology to summarize patent content in terms of claims, topics, functions, and technologies. Then, to classify the groups based on certain attributes, we introduce a clustering function. Lastly, the platform will make visualizations, which are used to interpret the patent where the technology or business trends and relations are predicted.

In terms of the patent analysis process, there are three stages: pre-processing stage, processing stage and post-processing stage. First, the patent data are collected, cleaned, and prepared comprehensively and the information is extracted during the pre-processing stage. Additionally, the extracted data are analyzed in the processing stage to classify, cluster, and determine valuable insight from the information. Those data are also analyzed using different methods. Finally, the post-processing stage is known as the process of knowledge discovery, where the result is visualized and evaluated so that we can make reasonable strategic decisions.

---

[#] These authors contributed equally to this work.

To develop the platform as stated above, the first part is collecting and training data. We collect raw data from the three data sources: Hong Kong Intellectual Property Department, Macau Economic and Technological Development Bureau, and China National Intellectual Property Administration. The platform will download the public data from the websites in the form of PDF. Next, the helpful information will be extracted and sent to our integrated database. In our simulation, our database has 36,786,954 pieces of patent data from mainland China, ranging from 2017/04/10 to 2022/04/10. There are also 117504 and 5745 pieces of patent data from Hong Kong and Macau, respectively, ranging from 2012/01/01 to 2022/04/10. We also present the sample patent data in our slides, and they are mainly presented in three languages: simple Chinese, traditional Chinese and English. Since enough data is collected, we make data classification in the next step. According to IPC, the hierarchical patent classification system consists of 8 sections, 128 classes, and 648 subclasses. Support Vector Machine (SVM) Genetic Algorithm is used to classify patent data into the sections or classes mentioned above.

Then, the system can start to make a valuation of each patent. There are two methods: the bibliographic approach and the content-based approach. For the bibliographic approach, there are some factors that influence market value, including citations, applicants, inventors, and IPC (International Patent Classification) codes. For the content-based approach, the factors, such as the technology involved, the pattern, trends, and opportunities extracted from the abstract, the summary, other detailed descriptions of the invention, and claims, will affect market value. In order to make the whole valuation process clearer, the flow chart is demonstrated in the slides. In the valuation process, the Neural Network is applied. In the first step, the system extracted useful data from MySQL database as our training data. Second, data is used to train the deep neural networks model. The output value is the market value of each patent we want to predict, presented in dollar amount units. In the third step, the performance of the network is tested. Next, input a random patent by our customers. Finally, output the predicted market value of the patent.

In addition to the market valuation, there is the patent quality evaluation. Similarly, there is a flow chart demonstrating in the slides. Before quality evaluation, the quality is defined using quality indicators, including legal events related to a patent, number of patent family, number of countries where a patent family is granted, number of citations the patent has received and number of non-patents who cite the patent. After we reduce the data dimension of the quality indicators using SOMs as stated above, we go through kernel principal component analysis (KPCA) for feature extraction, and finally use SVM to separate the quality of each level from 1 to 10. During the quality evaluation, characteristics will help the system judge the quality of patents. For characteristics, number of other patents that the patent cited, number of patent owners, number of technical areas the patent protected, number of classes in the patent, number of inventors, number of priority countries advocated in the patent and duration between the application date and approval date, are listed.

Machine learning algorithm defines a parameterized mapping function and an optimization algorithm is used to find the values of the parameters that minimize the error of the function when used to map inputs to outputs. In this system, Loss function and Gradient descent are applied to optimize the system. After linearization, the output of the value regression model will be the market value, measured by dollars. After linearization, the output of the quality classification model will be the ranking of quality, where the highest score is 10, and the lowest score is 1.

Finally, the task of the system is visualization. A topic map is used to present the result, and the python program is used to make visualizations. The main purpose of this part is to interpret the patent, predict technology, business trends and relations and make topical analysis.

### B. Market Valuation

The template is used to format your paper and style the text. All margins, column widths, line spaces, and text fonts are prescribed; please do not alter them. You may note peculiarities. For example, the head margin in this template measures proportionately more than is customary. This measurement and others are deliberate, using specifications that anticipate your paper as one part of the entire proceedings, and not as an independent document. Please do not revise any of the current designations.

### C. Maintaining the Integrity of the Specifications

The template is used to format your paper and style the text. All margins, column widths, line spaces, and text fonts are prescribed; please do not alter them. You may note peculiarities. For example, the head margin in this template measures proportionately more than is customary. This measurement and others are deliberate, using specifications that anticipate your paper as one part of the entire proceedings, and not as an independent document. Please do not revise any of the current designations.

### D. Data Security in Machine Learning

Training machine learning models requires a lot of data, which is not only contributed by a single individual or organization. By sharing data to collaboratively train the models, we can unlock value and develop powerful models for various scenarios. At the same time, we recognize the need to protect personal confidentiality and privacy, and win and maintain the trust of those who use our products. Protecting the confidentiality of customer data is the core of our mission.

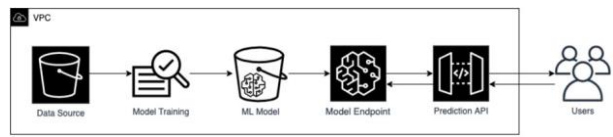

Figure 2 Example of a Basic Machine Learning Workflow

Figure 2 shows a basic ML workflow (Ng, 2021). To secure each stage of an ML workflow, from the data source to the prediction API, we will introduce several security measures.

1) Data encryption, which is the first line of defense to block unauthorized users from reading our data. We encrypt data both while it is in transit and at rest. A widely used cryptographic protocol is Transport Layer Security (TLS).

2) Differential privacy, which means adding small amounts of statistical noise during training to conceal the contributions of individual parties whose data is being used. When using DP, mathematical proof proves that the final ML model only learns general trends in the data and does not acquire information that is unique to any particular party.

3) Federated learning, which aims to keep raw training data remains within its local node, such as data silos or user devices, without any raw data leaving the node.

*E. Feasibility Analysis*

1) Commercial Feasibility

a) Benefit analysis

There are four main benefits to using ML to conduct our patent analysis, which is part of IP valuation.

Firstly, IP valuation methods traditionally rely on various economic valuation methods paired with expert analysis. But such methods, especially if directed at recent developments, can be somewhat speculative. The machine learning approach acts as a sword to cut down overly "hyped" IP and as a shield, protecting a valuable contribution even if fundamental in nature (early in research development). That implies far-reaching effects in transactional IP practice and along with the same reasoning, evaluation of IPOs. Beyond that added value, such automated practices may prove significantly less costly than identifying, hiring, and soliciting expert opinions.

Secondly, AI supervised learning (a machine learning algorithm) can help professional appraisers determine the value of patents more objectively and consistently. These improvements have increased confidence in the monetization of intellectual property transactions, which in turn has increased transaction flows in the financial system. By leveraging this new tool, a significant trend is emerging for commercial companies to use AI tools to pledge patents and monetize intellectual property more broadly.

Thirdly, when used in the proper context, our software programs can provide a less expensive, quicker, and more convenient method for evaluating certain aspects of a patent than retaining an expert to perform a manual analysis.

Moreover, Some traditional valuation institutions for intangible assets evaluate basic parameters such as earnings forecasts, divided into rate and discount rate value of the random and the lack of the basis, which leads to different appraisal institutions and even within the same appraisal institution personnel for the same appraisal object value conclusions far, minority appraisal institutions in order to occupy the market, use the ultra-low-cost contracting business, with the service advantage of less information and fast process, the report is submitted at an abnormal speed, resulting in a bad competitive environment for the evaluation of intellectual property pledge financing (Shan, 2017). Therefore, our software can solve problems like inconsistency and integrity.

Last but not least, we own a higher ability for data analysis, which is a great advantage compared to manual analysis.

b) Cost analysis

Our cost structure is mainly divided into development, maintenance, and equipment. For development, we need to spend funds in four areas: security, application, model, and data. Therefore we need to hire human forces from the different areas: risk management advisors, application development engineers, algorithm engineers, big data analysts, and data virtualization engineers. For maintenance, it means maintaining a good relationship with vital clients, so we need to hire a public relationship developer and maintainer. For equipment, it contains all kinds of devices for writing programs.

c) Competitive analysis

We are competing with a variety of big companies. They provide high-quality intellectual property valuation service and approach the valuation from many different angles supported by thorough research, financial modeling, and transparency of methodology. They have their own advantages and below are a few typical examples.

First of all, the industry leaders are the Big Four. They provide all types of valuation, including intangible assets. They adopt a range of complex valuation methodologies to produce robust analysis and bring together professionals with different kinds of expertise.

Black Stone IP, LLC is a boutique investment bank focused on valuing and trading tech and IP assets. Black Stone has more than 30 professionals who deploy deep market, financial, and IP knowledge to create brilliant value for clients.

Alix Partners' is a consulting firm. Their experts have significant experience in valuing intellectual property such as patents, trade names, trademarks, technology, in-process research and development, and other intangible assets such as customer relationships, license agreements, assembled workforces, and non-compete agreements.

FTI Consulting combines deep financial, forensic, economic, technology, and communications experience with extensive industry expertise worldwide. It helps clients accurately value intangible assets, including patents, trademarks, copyrights, and trade secrets. Intellectual property experts in the FTI Consulting Economics practice combine financial, econometric, and choice modeling to ascertain and present the value of IP assets to customers, licensees and regulators.

Above are the representative competitors that our company is facing, they have their advantages. At the same time, our company's advantages are different, which distinct our company.

2) Technical Feasibility

NLP is the technology that enables the machine to learn about contents in patents. Generally, there are three main processes of NLP, which are tokenization, pre-processing, and feature engineering. During tokenization, the actual words in the text are split from each other and then transformed into tokens for further processing. Then, tenses and plural forms of words are removed for English patent contents during pre-processing. After that, a word list is extracted from the whole text data during feature selection, so that we can compute the significance of every word in any specific patent, using the TF-IDF formula. By labelling the patents with their known value, we can compute each word's contribution to the average final value of the patent to build our final model.

3) Legal feasibility

In recent years, the field of IP appraisal utilizing machine learning has grown quickly. Because it is such a new technology, governments around the world have been slower to react to it. In the Greater Bay Area, it has not been the law until today. However, some countries and regions, such as Japan and the United States, show a strong and promising positive trend.

Japan Patent Office, has "automated patent-literature reviews, developed search algorithms to identify similar prior art, and automated classification of patent application by fields" (Ebrahim, 2018).

As what USPTO believes, automation technology and predictive analysis technology transform the pure human-to-human interaction between inventors and patent examiners into the interaction involving machine assistance, and "artificial-intelligence technology significantly outperforms humans and traditional statistical techniques between negotiation in the inventor–examiner interaction" (Ebrahim, 2018).

*F. Potential problems and solutions*

While developing a faultless IP evaluation solution, we also design coping ways to address any potential issues that may arise.

1) Data validity

It is possible that getting the data we want is difficult. There can be chances that the training data are of poor quality, and of low representativeness, resulting in over-fitting, under-fitting, weakened generalization ability, and insufficient accuracy. The key theme of our method is to increase the amount of training data, i.e., the number of patents. In this way, the non-conforming data will be reduced. As shown previously, the mainland China patent dataset already had almost 36 million patents. Apart from that, we use two other sources for our training data:

a) Patent data from our customers in the past. This is not going to work in the early stages of our development. However, once we have a fair volume of business, our consumers' recourse will be quite valuable.

b) Data exchange with other businesses in the same industry.

In the end, both of the above constitute data sharing. The entire procedure, from informed consent to post-use data processing, will be conducted in strict accordance with the regulations.

2) Prediction accuracy

It is not enough to have data alone. We also need human assessments for the success of our approach, particularly for accuracy validation. As a result, we plan to seek the assistance of IP valuation experts.

## III. BUSINESS IMPLEMENTATION

*A. Project Plan*

Our project plan is divided into four stages.

The first stage is "idea generation", during which we will come up with the ideas for our IP valuation software and corresponding ML algorithm. This stage will take us fifteen days. The second stage is "development". Implementing the idea from aspects like recruiting, supplier management, and software development will take us roughly five months. The third stage is the "marketing test". We will provide a private introduction to some target banks and persuade them to adoptions of our software and further improve our model. The last stage is commercialization. We will enhance our platform according to all channels of feedback. After improvements, we release our software to all potential customers.

*B. Financial Plan*

We divide the financing into three stages according to the development route of the enterprise. The first is the seed stage, where the valuation model is started to develop and tested. What we need are ideas and a team, the capital required is small so the funding source is our own resources. The second is the start and growth stage, where the model is further improved and we start trial sales. What we need is the professional database, office space, and staff salaries, the capital required increases significantly, and we could raise funds in the form of loans or shares from venture capital firms or angel funds. The third is the mature stage, where we will further expand the market with mature technology and considerable profit. The required funds are mainly used for market operations. With considerable return and mature technology, we could raise funds from private equity or the public market.

*C. Marketing Plan*

Our marketing objectives can be presented in three stages. First, for short-term objectives, we hope to tap the market of Intelligent patent evaluation platform in the Greater Bay Area and raise the reputation, within three months. Second, for mid-term objectives, we aim to hold 50% or more of the market share in the Greater Bay Area, rank among the top 5 in the domestic industry and develop international customers within 5 years. Third, for long-term objectives, in the future 15 years, we hope the company could be ranked among the top 50 in the industry. Moreover, our company can be successfully listed on the Hong Kong Stock Exchange.

There are four sections in the marketing plan. For marketing purposes, we will expand the market with a strong advertising campaign, accurately locate products, highlight product characteristics, and adopt differentiated marketing strategies. In addition, we will focus on the main consumer groups of products and expand the sales area. There are four subsections in the product strategy. First, our product positioning is for customers who need to evaluate the value of patents quickly with a low price. Second, to maintain the product quality and function, we will continuously train more data and optimize the model. In addition to patent valuation, we will develop valuation functions for all intellectual property. Third, for a better product brand, the reputation and popularity of our platform must be formed. Finally, we will improve customers' experience of the platform, and the quality of customer service. For price strategy, at the beginning, we give customers a free trial. After attracting more customers, we take the prime cost as the basis and the price of similar products as the reference. For advertising, we aim to establish the product image and company image, grasp the opportunity to carry out public relations activities and develop consumers. Additionally, actively use the social media and news to improve the popularity of enterprise products.